%% file: main.tex
\documentclass[10pt,twocolumn]{ICCAS}

\usepackage{eucal}
 
\usepackage{diagbox}

\usepackage{tcolorbox}
\usepackage{courier}
\usepackage{amsfonts}
 \usepackage{hyperref}
 \hypersetup{hidelinks}
\tcbuselibrary{skins, breakable}

\definecolor{promptheader}{RGB}{70, 130, 180}
\definecolor{promptborder}{RGB}{100, 160, 200}
\definecolor{promptbg}{RGB}{250, 252, 255}

\begin{document}

\title{Training-free Suction Grasp Detection for Deformed Aseptic Cartons Using Vision-Language Models and Geometric Surface Scoring}

\author{M. Maletić${}^{*}$ G. Vasiljević${}$ }

\affils{ Laboratory for Robotics and Intelligent Control Systems \\
University of Zagreb Faculty of Electrical Engineering and Computing, Unska 3, 10000 Zagreb, Croatia \\
\{marin.maletic, goran.vasiljevic\}@fer.unizg.hr \\ {\small${}^{*}$ Corresponding author}}


\abstract{
Robotic sorting of recyclable waste is challenging due to the deformable and geometrically inconsistent nature of target objects. 
We present a training-free suction grasping system for sorting deformed aseptic beverage cartons, decoupling target identification from grasp-point selection. 
An open-vocabulary vision-language model detects cartons from a text prompt, SAM2 refines each detection into an instance mask, and a geometric scoring method selects the suction point by combining surface flatness with normal alignment. 
Three geometric methods are compared: k-nearest-neighbour PCA, Sobel cross-product, and RANSAC plane fitting. 
Evaluated on a real robot across three deformation levels and 35 cluttered scenes, single-object grasp success reaches 88.2\% and end-to-end retrieval in clutter is 72.6\%.
}


\maketitle


\input{sections/1introduction}
\input{sections/2related_work}
\input{sections/3methods}

\input{sections/4experiments_results}

\input{sections/5conclusion}

\section*{ACKNOWLEDGEMENT}

This work was supported by the project Autonomous Robotic Technology for Environmental Monitoring, Intervention and Safety (ARTEMIS) DIGIT.2.1.02.14 financed through the Loan No. 9558-HR between the Republic of Croatia and the International Bank for Reconstruction and Development, which is part of the World Bank Group for the Digital, Innovation, and Green Technology Project (DIGIT Project).
The work of doctoral student Marin Maletić has been supported in part by the “Young researchers’ career development project training of doctoral students” of the Croatian Science Foundation.

%

%

\end{document}

%% file: sections/1introduction.tex
\section{Introduction}

Automated packaging sorting remains a major bottleneck in modern material recovery facilities (MRFs). 
Aseptic beverage cartons, commonly referred to as Tetra Pak packaging, present a particularly challenging manipulation target. 
In practical operating conditions, these cartons typically appear crushed, folded, and partially occluded on conveyor belts, resulting in geometries that differ substantially from the undeformed carton shapes on which conventional vision systems are generally trained. 
Vacuum suction is a preferred grasping modality for such lightweight objects, as a single airtight contact region is sufficient for reliable lifting. 
However, suction-based manipulation is highly sensitive to local surface characteristics: successful sealing requires a region that is sufficiently planar, non-porous, and oriented within the allowable tilt tolerance of the gripper; otherwise, grasp failure and object dropping are likely to occur.

Existing waste image datasets such as TACO \cite{taco} and ZeroWaste \cite{zerowaste} provide 2-D annotations for classification or segmentation but contain no depth data or grasp labels, and none isolate aseptic cartons in deformed states.

Over the past decade, supervised deep learning approaches have dominated suction grasp detection research. 
Existing methods typically learn grasp-quality functions from large-scale synthetic or manually annotated datasets and have demonstrated strong performance on warehouse inventory, household, and parcel-handling tasks. 
While these approaches currently define the state of the art, they are mostly trained on rigid or semi-rigid objects and treat each new object category as a distinct domain requiring dedicated training data. 
Furthermore, they lack the ability to adapt to new target classes without retraining. 
These assumptions are poorly suited to crushed and deformable beverage cartons, which violate both the geometric consistency and domain-specific training assumptions underlying existing methods.
\begin{figure}[htbp]
    \centering
    \includegraphics[width=7.5cm]{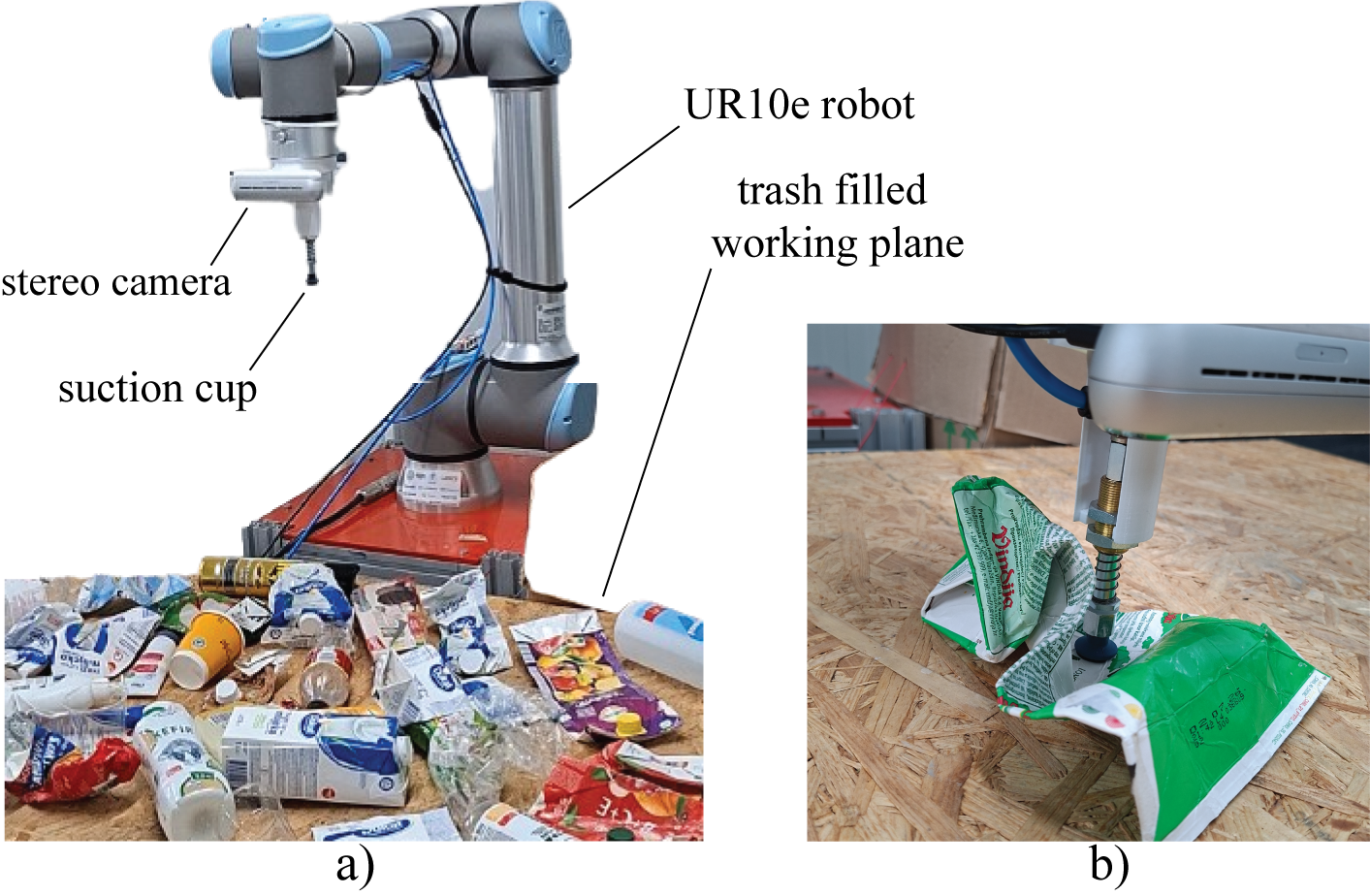}
    \caption{Experimental setup: (a) robotic workcell (b) successful suction grasp of heavily deformed carton}
    \label{fig:setup}
\end{figure}

This paper presents a training-free suction grasping framework for the automated sorting of deformed beverage cartons. 
The proposed system explicitly decouples target identification from grasp-point selection. 
First, a zero-shot vision-language model (VLM) identifies target objects from a free-form textual prompt. 
Subsequently, SAM2 \cite{sam2} generates refined instance segmentation masks for each detection. 
The segmented depth data are then back-projected to construct object-specific point clouds, after which a geometric evaluation stage ranks candidate suction points according to local surface flatness and surface normal orientation. 
Importantly, none of the system components are trained on the target domain, and the target object category can be modified dynamically at runtime through text prompting alone.

The contributions of this paper are: (i) a training-free pipeline integrating open-vocabulary VLM detection, SAM2 segmentation, and geometric suction scoring for robotic waste sorting; (ii) demonstrated grasping of deformed aseptic cartons, a class absent from existing suction benchmarks; and (iii) a comparative evaluation of three geometric scoring methods across deformation levels and clutter, including accuracy and runtime analysis.

%% file: sections/2related_work.tex
\section{Related work}
This section reviews three areas relevant to the proposed system: learned suction grasp detection methods and their limitations, geometric approaches to surface quality assessment, and recent open-vocabulary and VLM-guided grasping frameworks.

\subsection{Learned suction grasp detection}

Suction is favoured in industrial picking because one sealed contact often suffices to lift an object. 
Dex-Net 3.0 \cite{dexnet3.0} formalised this with a compliant contact model evaluating seal formation and wrench resistance, trained on 2.8 million synthetic point clouds while SuctionNet-1Billion \cite{Cao_2021} added a large-scale RGB-D benchmark and a learned predictor that surpasses it in clutter. 
Sim-Suction \cite{simsuction} released a 3.2-million-grasp synthetic dataset, Diffusion-Suction \cite{diffsuction} cast suction prediction as a denoising process with state-of-the-art results on a parcel dataset and OptiGrasp \cite{optigrasp} removed the depth sensor by adapting frozen foundation-model features to RGB. 
Each defines part of the current frontier, yet all train on rigid or semi-rigid distributions and treat every new object family as a separate domain requiring its own dataset, with no mechanism to retarget without retraining. Neither condition holds for crushed cartons.


\subsection{Geometric suction scoring}
Before learning-based methods, suction grasp candidates were selected using local surface normals and planarity, properties that remain central to physics-based grasp models. 
Random Sample Consensus (RANSAC) based plane fitting \cite{ransac} is widely used to separate support surfaces from objects. 
Lim \textit{et al.} \cite{lim2024} extended this to cluttered scenes using Multi-Object RANSAC and demonstrated competitive suction-picking performance. 
These results suggest that geometric reasoning remains effective when reliable object segmentation is available.

\subsection{Open-vocabulary and VLM-guided grasping}
Recent vision foundation models have significantly improved open-set perception and segmentation capabilities. 
SAM2 \cite{sam2} enables promptable instance segmentation, while Grounding DINO \cite{dino} performs open-vocabulary object detection from text queries. 
Several recent works, including GraspSAM \cite{noh2024}, HiFi-CS \cite{bhat2024}, and AffordGrasp \cite{tang2025} extend language grounding to robotic grasping, although they focus primarily on parallel-jaw grasping rather than suction manipulation. 
In parallel, multimodal systems such as Google DeepMind's Gemini Robotics \cite{geminirobotics2025} have demonstrated increasingly capable vision-language driven manipulation.

To the best of our knowledge, no prior work combines VL-based detection and SAM2 segmentation with geometric suction scoring for grasp selection. 
Consequently, suction grasping of deformable beverage cartons remains largely unaddressed in the existing literature.

%% file: sections/3methods.tex
\section{Methods}

\begin{figure*}[htbp]
    \centering
    \includegraphics[width=\textwidth]{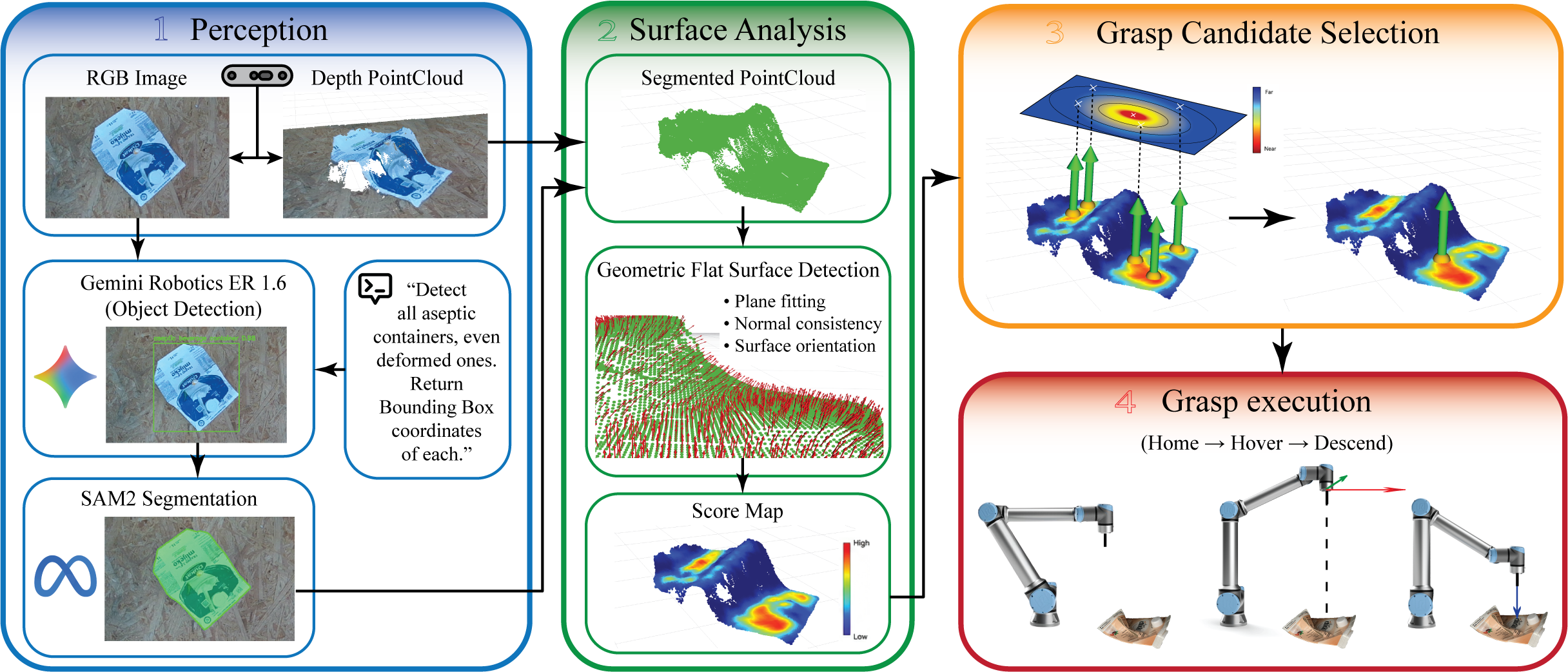}
    \caption{
    Overview of the pipeline. 
    \textbf{Perception:} The depth camera's RGB image is passed with a prompt to the VLM and the resulting bounding boxes are forwarded to SAM2 for segmentation. 
    \textbf{Surface analysis:} Depth pixels inside the mask are back-projected into a segmented point cloud, on which surface normals (red) and local flatness are estimated by one of three geometric methods, yielding a per-point suction quality heatmap. 
    \textbf{Grasp candidate selection:} among points clearing the sealing threshold, the one closest to the object centroid is chosen to minimise lift torque.  
    \textbf{Grasp execution:} The resulting pose is sent to the 
    robot arm, which approaches vertically, forms a vacuum seal, and lifts the carton.
}
    \label{fig:overal_shema}
\end{figure*}

The pipeline, illustrated in Fig \ref{fig:overal_shema}, consists of four stages, separating what to grasp from where on it to grasp: (i) a perception stage for open-vocabulary object detection and instance segmentation, (ii) a surface analysis stage for evaluating suction grasp suitability, (iii) a 3-D grasp selection stage for identifying the optimal grasp candidate and (iv) suction grasping execution.


\subsection{Perception}

The open-vocabulary reasoning capabilities of vision–language models allow our perception stage to detect cartons across different orientations and deformation states by reasoning about semantic appearance rather than matching to a fixed visual template.

The VLM is queried with the RGB camera frame $\mathbf{I} \in \mathbb{R}$$^{H \times W \times 3}$ alongside a structured natural-language prompt T.
The prompt specifies the task and requests a JSON array of bounding box pixel coordinates, as seen in Fig. \ref{fig:overal_shema}.
Each bounding box is passed as a spatial prompt to SAM2, which produces a binary instance mask. 
The valid depth pixels within each mask are back-projected to metric 3-D coordinates using the depth map $\mathbf{Z} \in \mathbb{R}$$^{H \times W}$ and the camera intrinsic matrix $K$, yielding a per-object labelled point cloud $\mathbf{P_k}$. 
Each point carries its instance index, its RGB image pixel coordinates (u, v), and its 3-D position $p=[x,y,z]$ in the camera frame.
All subsequent geometric operations are computed independently on each object's point cloud, resolving inter-object occlusion before suction scoring begins.

\subsection{Surface analysis}

For each detected object, every point $p=[x,y,z]$ in its point cloud is assigned a scalar suction quality score $s \in [0,1]$  formed as the product of two independent factors:
\begin{equation}
    s(\mathbf{p}) = f(\mathbf{p}) \cdot g(\mathbf{p}),
    \label{eq:score}
\end{equation}

\noindent where $f(\mathbf{p})$ is the \textit{flatness} term, quantifying how locally planar the surface is within the cup footprint, and $g(\mathbf{p})$ is the \textit{tilt-feasibility} term, quantifying how well the surface normal aligns with the gripper approach 
direction. 
A high score requires both conditions simultaneously: a flat but misaligned surface and a well-oriented but crumpled surface are each penalised. 

The gripper approach vector $\mathbf{a}$ is fixed perpendicular to the conveyor plane, matching the kinematic constraint of 3-DOF planar manipulators common in MRFs and exploiting the fact that downward contact pressure against the belt surface improves seal formation on deformable objects.
The surface tilt $\theta$ at point $\mathbf{p}$ relative to the gripper approach direction is:
\begin{equation}
    \theta(\mathbf{p}) = \arccos\bigl(\lvert \mathbf{n}(\mathbf{p}) 
    \cdot \mathbf{a} \rvert\bigr),
    \label{eq:tilt}
\end{equation}

\noindent where $\mathbf{n}(\mathbf{p})$ is the estimated surface normal at point $\mathbf{p}$ and $\mathbf{a}$ is the unit approach vector. 
Rather than applying a hard cutoff at the cup's maximum tilt tolerance $\theta_{\max}$, we define a transition band of half-width $\beta$ centered on $\theta_{\max}$:

\begin{equation}
    g(\mathbf{p}) = 1 -\text{clip}\!\left(\frac{\theta(\mathbf{p}) - (\theta_{\max} - \beta)}{2\beta},\ 0,\ 1\right),
    \label{eq:t}
\end{equation}

\noindent where $\text{clip}(\cdot,\, 0,\, 1)$ constrains the normalised displacement within the transition band to [0,1].
Therefore, the tilt-feasibility score $g(\mathbf{p}) \in [0,1]$ decreases linearly from 1 at $\theta_{\max} - \beta$ to 0 at $\theta_{\max} + \beta$. 

The three scoring methods described below differ only in how the surface normals $\mathbf{n}$ and flatness $f(\mathbf{p})$ are computed; the combination rule~(\ref{eq:score}) and the tilt term (\ref{eq:tilt}) are shared across all three.

\vspace{1mm}
\noindent\textbf{KNN method}

Following Hoppe et al. \cite{hoppe1992}, the surface normal at each point is estimated by principal component analysis (PCA) of its $k$ nearest 3-D neighbours, located via a k-d tree. 
For point $\mathbf{p}_i$ with neighbourhood $\mathcal{N}_{\text{i}}$, the local covariance matrix is:
\begin{equation}
    \mathbf{C}_i = \sum_{j \in \mathcal{N}_{\text{i}}}
    (\mathbf{p}_j - \mathbf{p}_i)(\mathbf{p}_j - \mathbf{p}_i)^\top.
    \label{eq:cov}
\end{equation}

\noindent We then perform eigendecomposition of $\mathbf{C}_i$ and extract $\mathbf{n}_i$ as the eigenvector corresponding to the smallest eigenvalue, which we use as the point normal. 
The normals are projected into an image-space normal map, and flatness is measured as the local consistency of those normals within a square window sized to the cup footprint in pixels. 
Let $\sigma(\mathbf{p})$ denote the mean per-channel standard deviation of the unit normals within the window centered at $\mathbf{p}$.
\begin{equation}
    f_{\mathrm{KNN}}(\mathbf{p}) = 
    \mathrm{clip}\!\left(1 - \frac{\sigma(\mathbf{p})}{\sigma_{\mathrm{tol}}},
    \; 0,\; 1\right),
    \label{eq:fknn}
\end{equation}

\noindent where $\sigma_{\mathrm{tol}}$ is the normal-variation level at which a surface is considered unsealable.

\vspace{1mm}
\noindent\textbf{Sobel method}

Following Nakagawa et al. \cite{nakagawa2015}, this approach estimates normals analytically from the organised structure of the point cloud, avoiding the per-point neighbour search entirely. 
The point cloud is arranged as a 3-channel image $\mathbf{I_{pc}}(u,v) = (x, y, z)$, and Sobel derivative filters are applied along both image axes to yield tangent vectors $\mathbf{t}_u = \partial\mathbf{I_{pc}}/\partial u$ and $\mathbf{t}_v = \partial\mathbf{I_{pc}}/\partial v$. 
The surface normal follows from their cross product:

\begin{equation}
    \mathbf{n}(u,v) = \frac{\mathbf{t}_u \times \mathbf{t}_v}
    {\lVert \mathbf{t}_u \times \mathbf{t}_v \rVert}.
    \label{eq:sobel}
\end{equation}

\noindent Flatness is then computed from these normals using the same window-variance rule as (\ref{eq:fknn}). 
Because the normals are obtained via dense image convolutions, the Sobel method is substantially faster than the KNN one, at the cost of greater sensitivity to depth sensor noise.

\vspace{1mm}
\noindent\textbf{RANSAC method}

The third grasp-scoring approach evaluates surface flatness in absolute physical units rather than through normal consistency. 
For each point, neighbouring points within a sphere whose radius equals the suction cup radius are collected, and random point triplets are sampled to generate candidate plane hypotheses \cite{ransac}. 
The plane with the highest inlier count, defined as points lying within a distance threshold $d_{\mathrm{thr}}$ from the plane, is retained, and its normal is assigned as $\mathbf{n}(\mathbf{p})$.
Surface flatness is then quantified as the root-mean-square deviation of the inliers, expressed in millimetres and evaluated against a physical sealing tolerance $\tau_{\mathrm{seal}}$:

\begin{equation}
    f_{\mathrm{RANSAC}}(\mathbf{p}) = 
    \mathrm{clip}\!\left(1 - 
    \frac{\mathrm{RMS}(\mathbf{p})}{\tau_{\mathrm{seal}}},
    \; 0,\; 1\right).
    \label{eq:fransac}
\end{equation}

\noindent This formulation makes the flatness term directly interpretable as a physical sealing-quality margin: surfaces whose root-mean-square deviation reaches $\tau_{\mathrm{seal}}$ receive a score of zero, independent of surface orientation. 
In addition, a linear edge penalty is applied to points located within one suction-cup radius of the mask boundary, since edge grasps rarely produce a full seal.

\subsection{Grasp candidate selection}

Before selection, per-point scores are averaged over the cup footprint, so the chosen contact point reflects seal quality over the entire contact area rather than a single surface point. 
Among all points whose smoothed score exceeds a sealing threshold $s_{\min}$, the system selects not the highest-scoring point but the one with the smallest horizontal distance relative to the object centroid $\bar{\mathbf{p}}$:

\begin{equation}
    \mathbf{p}^{\star} = 
    \underset{\mathbf{p}\,:\,s(\mathbf{p})\,\geq\, s_{\min}}{\arg\min}
    \;\bigl\lVert
        (\mathbf{p} - \bar{\mathbf{p}}) - 
        \bigl[(\mathbf{p} - \bar{\mathbf{p}}) \cdot \mathbf{a}\bigr]\,\mathbf{a}
    \bigr\rVert.
    \label{eq:selection}
\end{equation}

\noindent Minimising this offset reduces gravitational torque during lift, improving hold stability on irregular cartons. 
If no point clears the threshold, the system falls back to the global score maximum. 
The selected point and the approach-aligned orientation together define a 6-DOF grasp pose, which is published to the arm for execution.

%% file: sections/4experiments_results.tex
\section{Experiments and results}

Three experiments are conducted. 
The first evaluates scoring consistency and runtime across deformation levels using repeated observations of a stationary scene. 
The second measures grasp success on cartons for all scoring methods and deformation levels. 
The third assesses performance in cluttered mixed-waste scenes using detection recall, grasp success, and end-to-end retrieval rate.

\subsection{Experimental setup}
The system is evaluated on a physical setup as seen in Fig. \ref{fig:setup} a). 
Although the method is intended for deployment on a high-throughput 3-DOF planar manipulator such as a delta robot, all experiments reported here use a UR10e 6-DOF arm as a practical testing platform. 
An RealSense D455f depth camera is mounted on the arm wrist. 
During evaluation, the arm holds a fixed home position while capturing so that the camera observes the workspace from a static top-down view.
Suction is provided by a pneumatic vacuum generator supplied at 4.5 bar, terminating in a 15 mm diameter silicone suction cup. 

The perception stage uses the Gemini-Robotics-ER 1.6 model, accessed through its public API, for open-vocabulary detection, followed by SAM2 for segmentation. 
The geometric scoring stage is executed locally on Intel Core Ultra 9 CPU and an NVIDIA RTX 5070 GPU.

\begin{figure}[htbp]
    \centering
    \includegraphics[width=\linewidth]{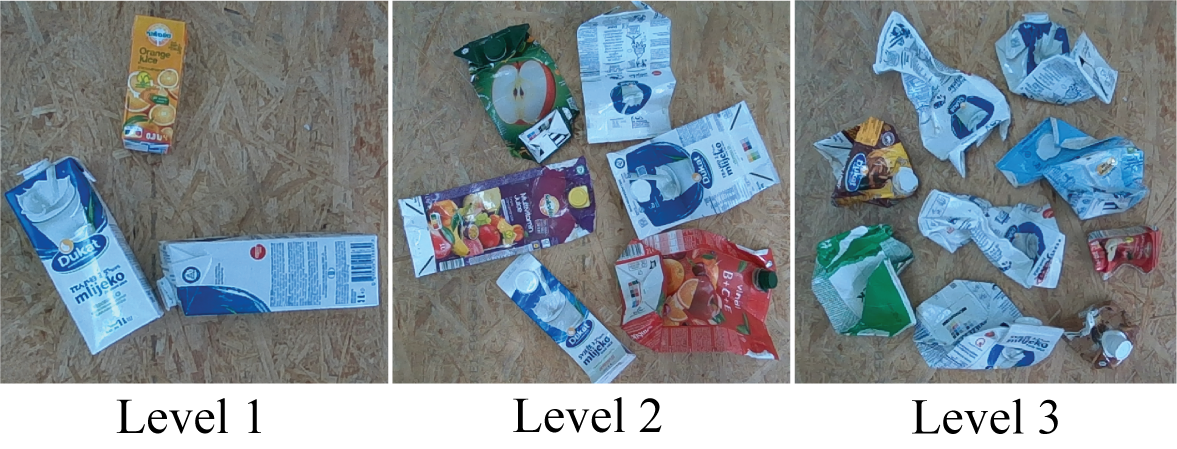}
    \caption{Examples of different carton deformation levels}
    \label{fig:levels}
\end{figure}

The evaluation uses a set of 40 commercial aseptic beverage cartons spanning a range of sizes, colours, and proportions. 
The set contains three levels of deformation: undeformed (Level 1), in which the carton retains its original prismatic shape; slightly deformed (Level 2), with mild folding and surface irregularity; and heavily deformed (Level 3), in which the carton is substantially crushed and its original geometry is largely lost. 
Examples of each level can be seen in Fig. \ref{fig:levels}. 
Each carton is evaluated across a range of orientations and poses, resulting in multiple trials per object. 

\subsection{Scoring Consistency and Runtime Analysis}
The first experiment isolates perception from robot execution and measures its stability under sensor noise. 
A single carton is held in a fixed pose while the detection–segmentation–scoring is run 15 times on independently captured frames. 
All three scoring methods are evaluated on the same detection output per frame.
A \textit{trial} is one (carton, pose) combination. 
Across the three levels, 84 trials and 1,491 frames were captured. 
The Gemini-Robotics-ER detector failed to recognise the carton in 1 frame at Level 2 and 18 frames at Level 3; these frames were excluded from the per-method analysis.

Table~\ref{tab:exp1} reports per-method consistency and runtime. 
All methods achieve millimetre-scale point spread, with means between 4.5 mm and 17.1 mm. 
However, per-trial maximum deviations reach 107–148 mm, where depth noise occasionally tips selection toward an alternative sealable region of comparable score.

\begin{table}[htbp]
\centering
\caption{Scoring consistency and runtime per method per deformation level. Consistency is the mean 3-D standard deviation (SD) of the selected suction point across 15 repeated runs, averaged over all $N$ trials (in mm). Scoring time is the per-frame computation time (in ms).}
\label{tab:exp1}
\renewcommand{\arraystretch}{1.15}
\begin{tabular}{clcccc}
\hline
\textbf{Lvl} & \textbf{Method} & \textbf{$N$} & \textbf{SD (mm)} & \textbf{Time (ms)} \\
\hline
1 & KNN    & 20 &  7.0 & $237 \pm 50$    \\
  & Sobel  & 20 & 11.9 & $\mathbf{127 \pm 6}$     \\
  & RANSAC & 20 &  \textbf{4.5} & $5610 \pm 3157$ \\
\hline
2 & KNN    & 30 & 12.4 & $266 \pm 65$    \\
  & Sobel  & 30 & 17.1 & $\mathbf{135 \pm 7}$     \\
  & RANSAC & 30 & \textbf{12.2} & $6105 \pm 3423$ \\
\hline
3 & KNN    & 34 &  \textbf{9.2} & $212 \pm 27$    \\
  & Sobel  & 34 & 11.9 & $\mathbf{137 \pm 10}$    \\
  & RANSAC & 34 &  9.8 & $2116 \pm 876$  \\
\hline
All & KNN    & 84 &  9.9 & $239 \pm 55$    \\
    & Sobel  & 84 & 13.8 & $\mathbf{134 \pm 9}$     \\
    & RANSAC & 84 &  \textbf{9.4} & $4477 \pm 3249$ \\
\hline
\end{tabular}
\end{table}

Runtime spans two orders of magnitude (Sobel 134 ms, KNN 239 ms, RANSAC 4477 ms). 
The VLM API call dominates at 5321 ms and SAM2 adds 145 ms, so the lighter scorers are negligible against perception while RANSAC roughly doubles pipeline time.


\subsection{Single-Object Grasp Success}

The second experiment evaluates end-to-end grasp execution on isolated cartons. 
For each carton, deformation level, and pose, the pipeline runs once per scoring method and the arm attempts the resulting grasp. 
Success requires the cup to form a seal and the carton to remain held through the full lifting motion to the drop-off area.

Table~\ref{tab:exp2}  reports per-level success rates with Wilson 95\% confidence intervals. 
All methods reach 100\% at Level 1. 
Performance degrades with deformation: at Level 3, rates range from 69.7\% (KNN) to 78.8\% (RANSAC). 
Pooled across all deformations, RANSAC and Sobel both reach 88.2\% and KNN 84.9\%.

Two categories of failure were observed:(i) detection misses by the VLM (3 at Level 2 and 5 at Level 3) and (ii) one large carton exceeding the lifting capacity of the 15 mm cup (the seal formed but broke during lift  due to gravitational torque).
On one Level 3 carton interesting scoring behaviour was observed when KNN and Sobel both succeeded but selected different surface regions.

\begin{table}[htbp]
\centering
\caption{Single-object grasp success rate per scoring method per deformation level. Shown as successes/attempts (rate). Brackets give Wilson 95\% confidence intervals.}
\label{tab:exp2}
\renewcommand{\arraystretch}{1.15}
\begin{tabular}{cccc}
\hline
\textbf{Lvl} & \textbf{KNN} & \textbf{Sobel} & \textbf{RANSAC} \\
\hline
1 & 30/30 (1.00) & 30/30 (1.00) & 30/30 (1.00) \\
  & [.89, 1.00]  & [.89, 1.00]  & [.89, 1.00]  \\
\hline
2 & 26/30 (.87)  & \textbf{27/30 (.90)}  & 26/30 (.87)  \\
  & [.70, .95]   & \textbf{[.74, .97]}   & [.70, .95]   \\
\hline
3 & 23/33 (.70)  & 25/33 (.76)  & \textbf{26/33 (.79)}  \\
  & [.53, .83]   & [.59, .87]   & \textbf{[.62, .89]}   \\
\hline
All & 79/93 (.85)  & \textbf{82/93 (.88)}  & \textbf{82/93 (.88)}  \\
    & [.76, .91]   & \textbf{[.80, .93]}   & \textbf{[.80, .93]}   \\
\hline
\end{tabular}
\end{table}


\subsection{Cluttered Scene Performance}

\begin{figure}[htbp]
    \centering
    \includegraphics[width=\linewidth]{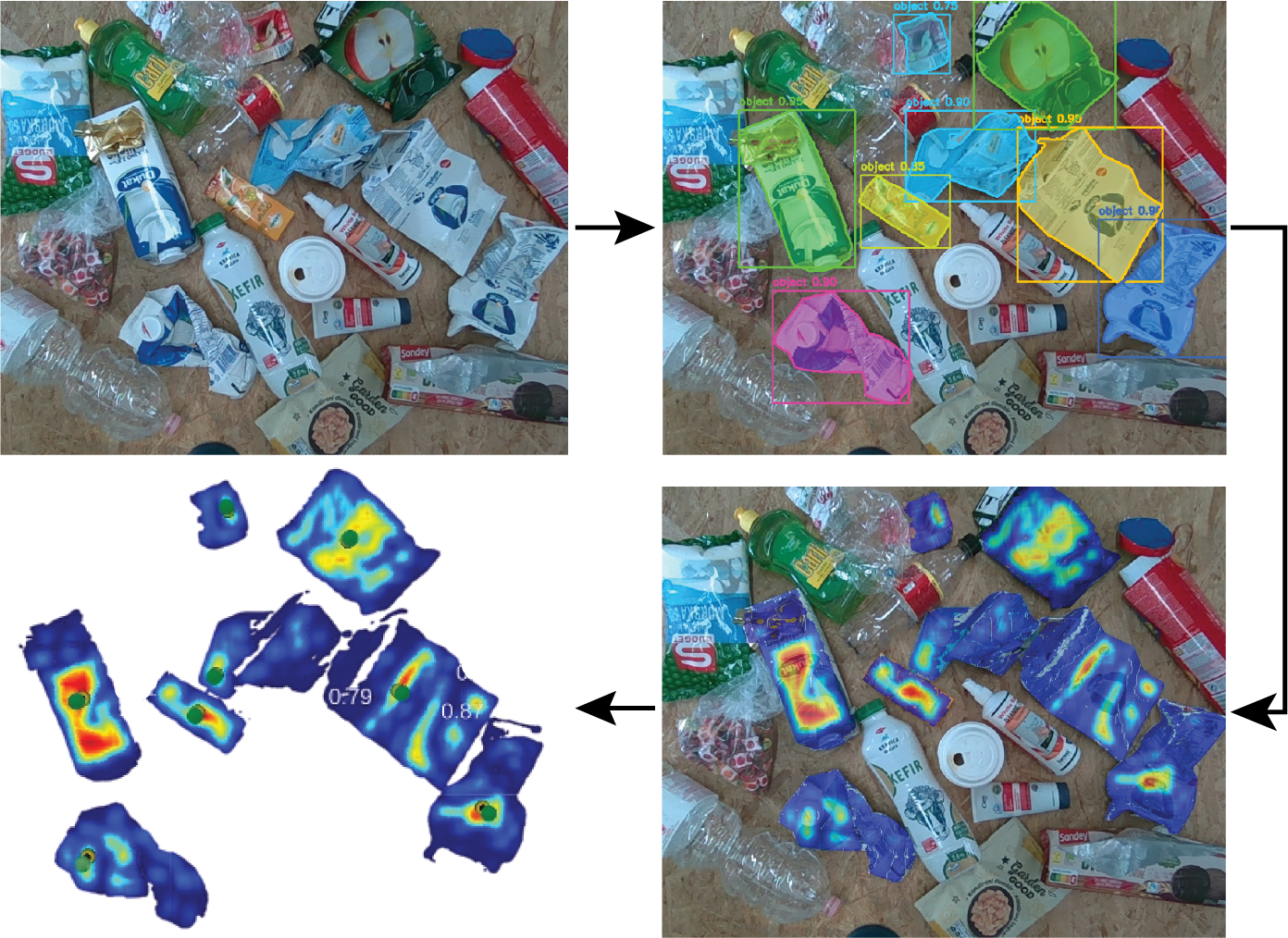}
    \caption{Exp. 3 - Detection and scoring visualized}
    \label{fig:clutter-scene}
\end{figure}

The third experiment evaluates the system under realistic sorting conditions. 
35 scenes were randomly constructed to emulate the operating environment of a MRF, with mixed plastic packaging waste — bottles, wrappings, and other containers typical of post-consumer recycling streams, placed alongside aseptic cartons. 
Scenes approximate the scenario of workers spreading collected bag contents onto a conveyor before robotic pickup: objects are intermixed but heavy occlusion and deep piles were avoided, with cartons mostly visible from above. 
The number of cartons per scene was varied from 2 to 8 containing all deformation levels, with 175 cartons present in total across the 35 scenes. 
For each scene, the system performed one detection–segmentation–scoring cycle in the beginning, and then sequential execution, attempting to pick each detected carton.
Based on the results of Experiment 2, the Sobel-based scoring method was selected for the final system, as it provided a balance between grasping success and computational efficiency.
Fig.~\ref{fig:clutter-scene} illustrates an example scene of cartons in clutter and their detection, segmentation, and scoring  with selected grasp points in green.

Table~\ref{tab:exp3} reports the three key metrics. 
The VLM detector recognised 146 of the 175 cartons present, yielding a pooled detection recall of 83.4\% (Wilson 95\% CI: 77.2–88.2\%). 
Of those 146 detected cartons, 127 were successfully grasped, for a conditional grasp success rate of 87.0\% (95\% CI: 80.6–91.5\%). 
The resulting end-to-end retrieval rate (cartons retrieved out of cartons present) is 72.6\% (95\% CI: 65.5–78.6\%).

\begin{table}[htbp]
\centering
\caption{Cluttered-scene performance across 35 scenes Sc. (15 sparse and 20 dense) containing 175 cartons total, mixed among distractors. Detection recall = detected/present; conditional grasp success = grasped/detected; end-to-end retrieval = grasped/present.}
\label{tab:exp3}
\renewcommand{\arraystretch}{1.2}
\begin{tabular}{cccccc}
\hline
\textbf{Cartons} & \textbf{Sc.} & \textbf{Det. recall} & \textbf{Grasp/Det.} & \textbf{E2E} \\
\hline
2-4 & 15 & 86.7\% & 92.3\% & 80.0\% \\
5-8  & 20 & 82.3\% & 85.0\% & 70.0\% \\
\hline
\textbf{175} & \textbf{35} & \textbf{83.4\%} & \textbf{87.0\%} & \textbf{72.6\%} \\
\hline
\end{tabular}
\end{table}

The end-to-end rate is bounded primarily by detection recall: of the 48 cartons not retrieved, 29 were missed by the detector and 19 failed at execution.
Detection recall and conditional grasp success both decrease as scene density increases (86.7\% $\xrightarrow{}$ 82.3\% and 92.3\% $\xrightarrow{}$ 85.0\%), consistent with greater object occlusions reducing VLM detection confidence and limiting valid grasp surfaces.

\begin{figure*}[t]
    \centering
    \includegraphics[width=\linewidth]{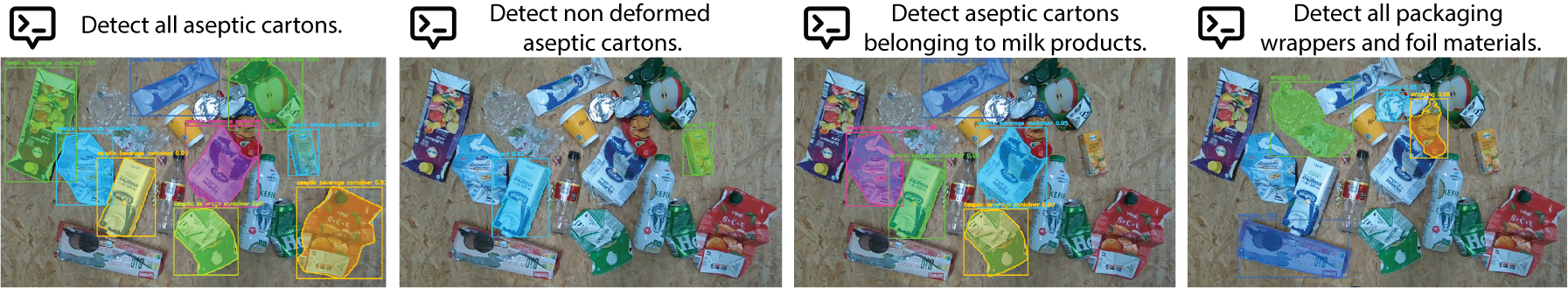}
    \caption{Prompt-driven target selection - detection and segmentation outputs for different natural-language prompts}
    \label{fig:prompt}
\end{figure*}

\subsection{Prompt-Driven Target Selection}
Because target identification is driven entirely by the text prompt, the set of objects the system attempts to grasp can be redefined at runtime without any retraining. 
Fig. \ref{fig:prompt} shows the same scene queried with four different prompts: all aseptic cartons, heavily deformed only, milk cartons only, and plastic wrappings, each producing a distinct detection and segmentation output while the geometric scoring stage remains unchanged. 
This runtime retargetability is structurally unavailable to supervised grasp detectors, which fix their target class at training time.

%% file: sections/5conclusion.tex
\section{Conclusion}

This paper presents a training-free suction grasping system for the automated sorting of deformed aseptic beverage cartons. 
By combining an open-vocabulary vision-language detector and SAM2 segmentation with purely geometric grasp-point scoring, the system avoids domain specific training entirely and supports run-time retargeting through natural language. 
Across three deformation levels and cluttered scenes containing mixed plastic waste, the system achieved 87.0\% conditional grasp success in clutter and 72.6\% end-to-end retrieval. 
Detection recall is the primary bottleneck, as 60\% of non-retrievals stemmed from VLM failures rather than execution errors. 
Future work will address re-planning after failed grasps, seal-quality feedback, and conveyor-speed throughput for industrial MRF deployment.

All accompanying code and the video of the pipeline in action is publicly available through the LARICS Lab GitHub: \href{https://github.com/larics/geoSuctionBot.git}{https://github.com/larics/geoSuctionBot.git}.